\def\year{2022}\relax
\documentclass[letterpaper]{article} 
\usepackage{aaai22}  
\usepackage{times}  
\usepackage{helvet}  
\usepackage{courier}  
\usepackage[hyphens]{url}  
\usepackage{graphicx} 
\usepackage{natbib}  
\usepackage{caption} 
\DeclareCaptionStyle{ruled}{labelfont=normalfont,labelsep=colon,strut=off} 
\usepackage{algorithm}
\usepackage{algorithmic}

\usepackage{cite}
\usepackage{amsmath,amssymb,amsfonts}
\usepackage{graphicx}
\usepackage{textcomp}

\usepackage{bm}
\usepackage{xcolor}
\usepackage{dsfont}
\usepackage{booktabs}
\usepackage{caption,subcaption}

\usepackage{newfloat}
\usepackage{listings}
\floatstyle{ruled}
\newfloat{listing}{tb}{lst}{}
\floatname{listing}{Listing}
\title{PRANet: Point Cloud Registration with an Artificial Agent}
\author{
      Lisa Tse\textsuperscript{\rm 1,3},
    Abdoul Aziz Amadou\textsuperscript{\rm 2,3},
    Axen Georget\textsuperscript{\rm 3},
    Ahmet Tuysuzoglu\textsuperscript{\rm 3}\\
}
\affiliations{
\textsuperscript{\rm 1}Department of Computer Science, University College London, England, UK\\
\textsuperscript{\rm 2}School of Biomedical Engineering \& Imaging Sciences, King's College London, England, UK\\
\textsuperscript{\rm 3}Siemens Healthineers,
    Digital Technology and Innovation,
    Princeton NJ, USA

}

\newcommand{\src}{\mathcal{X}}
\newcommand{\srci}{\bm{x}}
\newcommand{\bx}{\bm{x}}
\newcommand{\tgt}{\mathcal{Y}}
\newcommand{\tgti}{\bm{y}}

\newcommand{\trs}{\bm{T}}
\newcommand{\rot}{\bm{R}}
\newcommand{\SO}{\mathrm{SO}}
\newcommand{\SE}{\mathrm{SE}}
\newcommand{\trans}{\bm{t}}

\newcommand{\id}{\mathds{1}}
\newcommand{\tr}{\mathrm{tr}}
\DeclareMathOperator*{\argmax}{arg\,max}
\DeclareMathOperator*{\argmin}{arg\,min}

\newcommand\blfootnote[1]{%
  \begingroup
  \renewcommand\thefootnote{}\footnote{#1}%
  \addtocounter{footnote}{-1}%
  \endgroup
}

\begin{document}

\maketitle
\blfootnote{The concepts and information presented in this abstract/paper are based on research results that are not commercially available. Future availability cannot be guaranteed.}

\begin{abstract}
Point cloud registration plays a critical role in a multitude of computer vision tasks, such as pose estimation and 3D localization. Recently, a plethora of deep learning methods were formulated that aim to tackle this problem.
Most of these approaches find point or feature correspondences, from which the transformations are computed. We give a different perspective and frame the registration problem as a Markov Decision Process. Instead of directly searching for the transformation, the problem becomes one of finding a sequence of translation and rotation actions that is equivalent to this transformation. To this end, we propose an artificial agent trained end-to-end using deep supervised learning. In contrast to conventional reinforcement learning techniques, the observations are sampled i.i.d. and thus no experience replay buffer is required, resulting in a more streamlined training process. 
Experiments on ModelNet40 show results comparable or superior to the state of the art in the case of clean, noisy and partially visible datasets.
\end{abstract}
\section{Introduction}
\begin{figure}[ht]
\centering
  \begin{subfigure}[b]{\linewidth}
	\centering
	\includegraphics[width=0.53\linewidth,trim={2.5cm 2cm 4cm 2cm},clip]{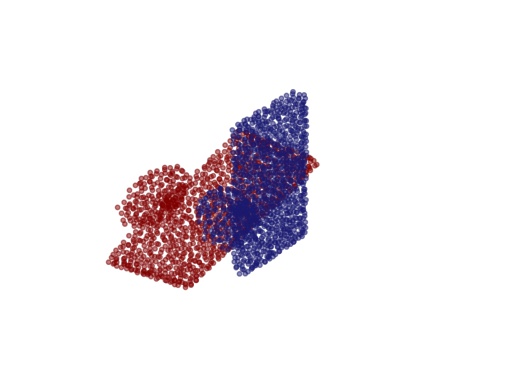}
	\includegraphics[width=0.46\linewidth,trim={3.5cm 2cm 4cm 2cm},clip]{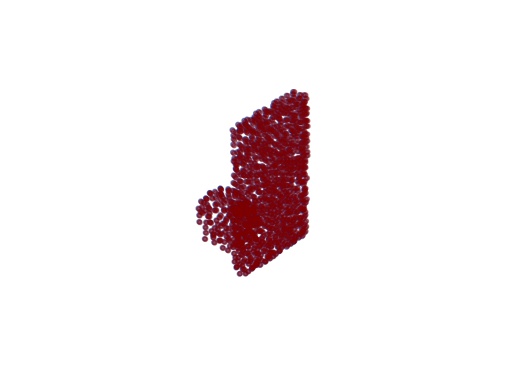}
  \end{subfigure}
\caption{Example point clouds before and after registration for the noiseless case. }
\label{fig:example_registration}
\end{figure}

Point cloud registration is the task of aligning two point clouds that differ from each other by a transformation, possibly made more challenging through the presence of noise, outliers and occlusions. In this work, we consider rigid transformations through rotations and translations. An example registration problem can be found in Fig~\ref{fig:example_registration}. The rigid registration problem has widespread applications in robotics, medical imaging and computer vision. The task becomes trivial once the correspondences between the points in the two point clouds are known, and this is a popular approach to solving the registration problem. However, finding correspondences is in itself a challenging problem, particularly in the presence of outliers and sampling differences meaning that some or all points do not have exact correspondences.

Recently, works such as~\cite{Aoki2019, Wang2019,Wang, Yew2020,idam} proposed learning-based methods that provided promising results. The vast majority of these approaches attempt to find point or feature correspondences and then apply a singular value decomposition (SVD) to obtain the transformation. We, instead, frame the point cloud registration problem as a Markov Decision Process (MDP) and solve it by learning reward functions using deep supervised learning, which does not require the prediction of correspondences. Unlike conventional reinforcement learning methods, the training samples are taken i.i.d. and thus no experience replay buffer is needed. The idea of using deep supervised learning was also used by~\cite{Miao2017}, which applied it to the problem of image registration. To the best of our knowledge, our work is the first to pose the point cloud registration problem as an iterative agent solving an MDP and trained using deep supervised learning. Our method does not rely on estimating correspondences and therefore encourages the use of a new paradigm to approach this problem. The work in~\cite{Bauer2021} also avoids point correspondences by using deep reinforcement learning. However, it requires an experience replay buffer and a full sampling along a registration trajectory, as well as the use of imitation learning. In contrast, our proposed method does not require the creation of replay buffers nor the use of imitation learning, hence resulting in lower memory requirements and an easier training process. 

\subsubsection{Contributions.} Our main contributions are as follows:
\begin{itemize}

\item We pose the point set registration problem as an MDP coupled with a reward function that measures the reduction of distance in the transformation space. 

\item We devise and train an artificial agent using deep supervised learning based on the MDP formulation, providing a novel approach without resorting to point correspondences.

\item We perform an unbiased sampling of the $\SO(3)$ manifold and conduct in-depth experiments to show the benefits of our approach in clean, noisy and partially visible datasets (trained on unseen categories).

\end{itemize}

\section{Related Work}

\subsubsection{Traditional registration methods.}

Iterative Closest Point (ICP) is possibly the most well-known approach to the point cloud registration problem. ICP iterates between estimating the rigid transformation and point correspondences. Given correspondences, finding the transformation reduces to an SVD-based operation. The performance of ICP largely hinges on the initialization and it can easily get trapped in suboptimal local minima depending on initial correspondences or transformation assumed. There are many variants such as \cite{Segal,Yang2016,Rusinkiewicz2001,Fitzgibbon2003} that aim to overcome the shortcomings of classical ICP. In particular, \cite{Yang2016} offers a global approach through a branch-and-bound scheme, although it is inefficient for practical purposes. 
Another global method optimizes a global objective on the correspondences between source and target point clouds~\cite{Zhou2016}. \cite{Yang2021} proposed a certifiable algorithm that provides high robustness to outliers using a truncated least-squares loss.

\subsubsection{Learning based registration methods.}
The possibility of encoding point clouds directly into deep neural networks was made possible by the pioneering work of PointNet~\cite{Qi2017}, which takes into account the permutation invariance of the points to produce a global embedding for a given point cloud.
Building on PointNet, \cite{Aoki2019} combines this global embedding with the classical, iterative Lucas-Kanade method~\cite{Lucas1981} to perform registration by aligning the global features of the transformed source and target point clouds. \cite{Wang} trains an end-to-end network by using the DGCNN embedding, proposed in~\cite{DGCNN}, with an attention mechanism, matching the point embeddings and then extracting the transformation from the matching matrix through a differentiable SVD layer. \cite{Wang2019} proposes a more robust algorithm to this and matches only select keypoints, showing improved experimental results for partial-to-partial registration. \cite{Yew2020} takes advantage of surface normals given in the data to perform registration based on the classical RPM method. The work of~\cite{Bauer2021} is similar to ours in that it is based on an artificial agent, but it requires imitation learning as part of its initialisation as well as an experience replay buffer. \cite{rgmnet} embeds each point cloud as a graph and performs embedding matching to obtain the transformation.  \cite{idam} has a matching process that takes into account both the embedding space and the Euclidean point space. In contrast to our work, the vast majority of the these methods perform some variant of point matching to extract the final transformation. An artificial agent similar to our approach is applied on image registration in~\cite{Miao2017}. Although they also learn a reward vector over a set of incremental actions to perform the registration, their embedding uses convolutional neural networks along with a different reward scheme that couples rotations and translations. 

\section{Problem Statement}
We are concerned with the problem of rigid point cloud registration. Given a source point cloud $\src =\{\srci_1,\ldots,\srci_K\}\subset \Re^{3}$ and target point cloud $\tgt=\{\tgti_1,\ldots,\tgti_L\} \subset \Re^{3}$, the aim is to find a rigid transformation $\trs = (\rot,\trans)$ that minimizes the following cost function:
\begin{eqnarray}
&\argmin_{\rot, \trans, \bm{M}}\| \rot\src + \trans\mathbf{\underline{1}}  -  \tgt \bm{M}\|_p^p& \\
&\text{ s. t. } \rot \in \SO(3)& \nonumber\\
&\bm{M}\mathbf{\underline{1}} \leq \mathbf{\underline{1}}, \,\,\,
\mathbf{\underline{1}}^T\bm{M} \leq \mathbf{\underline{1}}^T, \bm{M}_{i,j} \in \{0,1\}& \nonumber
\end{eqnarray}
where an $\ell_p$-norm distance of the registered and permuted point sets is minimized with respect the transformation parameters and the permutation matrix $\bm{M} \in \{0,1\}^{K\times L}$. The constraints on the permutation matrix allow for the existence of points in one set that do not correspond to a point in the other, such as in the presence of sampling differences and occlusions. The manifold constraint due to $\SO(3)$ for $\rot$ and the binary value constraint for $\bm{M}$ result in a non-convex problem that is challenging to solve. Most of the aforementioned learning based methods \cite{Wang,Wang2019,Yew2020} obtain an estimate of $\bm{M}$ and then solve the resulting easier problem via the use of polar decomposition.

In this work we take a different approach by using an MDP based formulation equipped with a reward function that measures the reduction of distance in the transformation space. We define the MDP by the tuple $\{S, A, p,r\}$, where $S$ denotes the set of states, $A$ the set of actions, $p_a(s,s')=P(s_{t+1}=s'|s_t=s,a_t=a)$ gives the transition probability of arriving at state $s'$ at time $t+1$ after taking action $a$ in state $s$ at time $t$, and $r_a(s)$ denotes the reward associated with taking the action $a$ in state $s$. For our problem, we take $S$ to be the set of all possible rigid transformations ($\SE(3)$), and $A$ is a finite set of rotation or translation actions. Concretely, the actions consist of positive and negative translations of fixed magnitudes in each of the $x$, $y$ and $z$ directions, as well as rotations of fixed angles around each of the three axes in the positive and negative directions. When an action $a$ is applied, the new state $s'$ is given by the composition of the transformations represented by the previous state $s$ and the action. This composition is deterministic, giving deterministic transitions so that $p_a(s,s')\in\{0,1\}$ for any states $s,s' \in S$. The result of the composition is only dependent on the previous state and independent of other past states, fulfilling the Markovian property of $P(s_{t+1} | s_{t},\ldots,s_1, a_t) = P(s_{t+1} | s_{t}, a_t)$ for all time steps $t$. 



The registration problem can then be solved by performing an optimal set of actions that maximize the reward. This is formalized by the policy $\pi$, which returns a probability distribution of actions given a state. We consider a deterministic and greedy policy, given by $\pi(s) = \argmax_a {r_a(s)}$.

\begin{figure*}
\centering
\includegraphics[width=\textwidth]{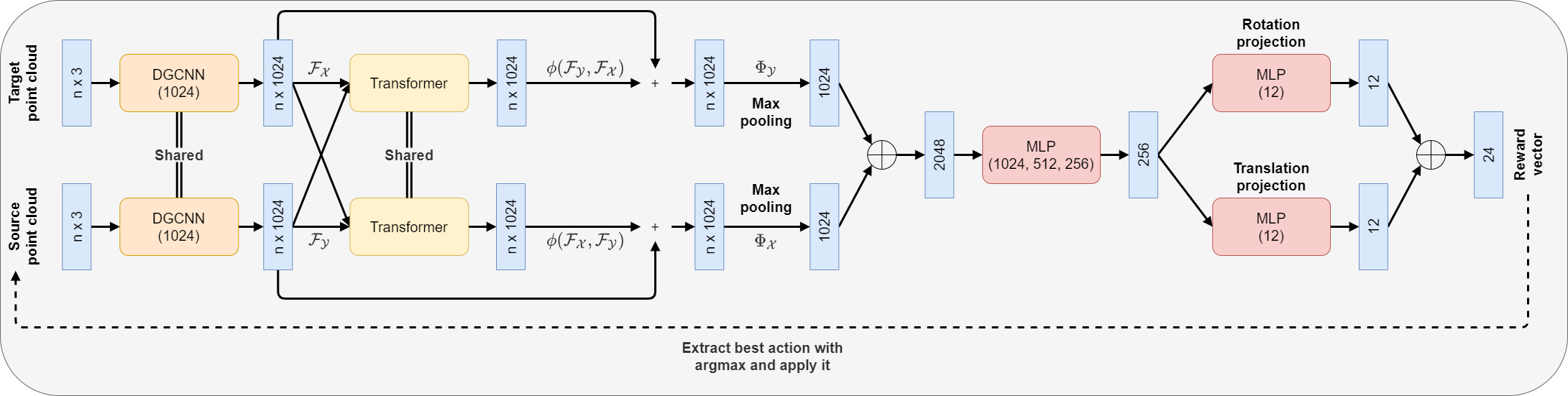}
\caption{Architecture of our method. The process marked by the dotted line is only performed during inference, while the other processes are done during both training and inference. The source and target point clouds share the weights for the DGCNN and Transformer modules.}
\label{fig:architecture}
\end{figure*}

\section{Artificial Agent}

We solve the MDP based point cloud registration problem  through a greedy approach inspired by $Q$-Learning. In this setup, we aim to learn the reward function by modelling it as a deep neural network. The policy we undertake is the action that maximizes the reward
function for a given state at time $t$, $\trs_t$: $a_{t} = \argmax_a r_a(\trs_t)$. To learn this, we sample source point clouds from the dataset and transformations from the $\SO(3)$ manifold uniformly at random. From these, we create the corresponding target point clouds, and feed the two point clouds into the neural network. The network then outputs a reward vector, where each element represents the reward for each action that is then regressed according to the ground truth rewards. 

The samples are taken i.i.d. and as such, we do not continue the registration trajectory with the policy during training time. This eliminates the need for a memory replay buffer, thereby reducing the memory requirements during training. Similarly, for the validation phase, we compute the loss of the reward vector instead of the registration error. At test time, we apply the action with the highest reward iteratively to the source point cloud. The reward is designed so that the source point cloud converges towards the target point cloud as the registration progresses. Thus, instead of implementing a stopping action, the registration process is allowed to run for a fixed number of steps.


\subsection{Reward function}
We chose a reward function that gives the reduction of distance to the identity transformation: \begin{equation}r_a(\trs) = D(\trs,\id) - D( a \oplus \trs,\id),\label{eq:reward}\end{equation} where $\oplus$ denotes the composition.  
The distance function is of the form \[D(\trs_1,\trs_2) = D_{\trans}(\trans_1,\trans_2) + D_{\rot}(\rot_1, \rot_2), \] where \[D_{\trans}(\trans_1,\trans_2) = ||\trans_1 - \trans_2 ||_2\] and \[D_{\rot} (\rot_1,\rot_2) = \arccos\left(\frac{\tr(\rot_1 \rot_2^{-1}) -1}{2} \right).\] It can be shown that $D_{\rot} (\rot_1,\rot_2)$ is equivalent to half the rotation angle of the combined rotation $\rot_1\rot_2$. It is a metric on $\SO(3)$ and it has a linear relationship with the geodesic on the sphere (see~\cite{Huynh2009} for a more in-depth discussion).  For actions that are purely rotational, the translational contributions to the two terms in~\eqref{eq:reward} are identical and therefore cancel out. The same holds for purely translational actions and rotational contributions. 

\subsection{Application of actions}
Since the contributions of purely rotational and translational actions are decoupled in our reward function, we apply these in a similarly decoupled fashion during inference time. This is particularly important for rotations, where the naive method would be to apply it directly to the point cloud at each iteration. However, this would induce an additional translation since the source point cloud is not centered at the origin in general. To see this, we start with a point cloud $\src$ which is centered around the origin and translate it by the vector $\trans$ . Applying an additional rotational action $\rot$ to the resulting point cloud would give
\[\rot \left (\src + \trans \right) = \rot \src + \rot \trans. \] 
Instead, we would like to obtain $\rot \src + \trans$ where the translation vector remains unmodified. For $T$ iterative rotations $(\rot_1,\ldots, \rot_T )$ and $S$ iterative translations $(\trans_1,\ldots,\trans_S),$ we thus combine these transformations as follows:
\[ \rot_T \ldots \rot_1 \src + \sum_{i=1}^S \trans_i. \]

\subsection{Architecture}
Fig.~\ref{fig:architecture} shows the architecture that we use. We use a standard embedding for the point clouds and then extract the rewards from this using a multilayer perceptron (MLP). The action step sizes are learned implicitly and we do not provide these to the network. Our embedding is inspired by that of DCP~\cite{Wang}, using DGCNN \cite{DGCNN} followed by a Transformer. DGCNN creates an embedding that aims to exploit local geometric features. It does so by constructing a dynamic $k$-NN graph, and applying convolution-like operations on the edges iteratively for multiple layers. At a layer $l$, the following update is performed for the index $m\in[M_l]$ of a point embedding $\bx_i$:
\[ \bx_{i,m}^l = \max_{j: (i,j)\in \mathcal{E}_{l-1} } h^m_{\bm{\Theta}_l}( \bx_{i}^{l-1},\bx_{j}^{l-1}) ,\]
where $\mathcal{E}_{l}$ is the set of edges in the graph at layer $l$, $\bm{\Theta}_l$ encodes the weights associated with layer $l$ and $h_{\bm{\Theta}_l}^m(\bx_i,\bx_j):=\mathrm{ReLu}(\bm{w}_{m,l} \cdot \bx_i  + \bm{v}_{m,l} \cdot  \bx_j)$ where $\bm{\Theta}_l=(\bm{w}_{1,l},\ldots,\bm{w}_{M_l,l}, \bm{v}_{1,l}, \ldots,\bm{v}_{M_l,l})$. Given a point cloud with $N$ points, DGCNN outputs an embedding in $\Re^{N\times K}$, where $K$ is the embedding dimension.

The Transformer module allows for an attention mechanism between the point clouds. Taking $\mathcal{F}_\src$ and $\mathcal{F}_\tgt$ to be the source and target embeddings generated by DGCNN respectively, the output source and target embeddings $\Phi_\src$ and $\Phi_\tgt$ after the Transformer are given by:
\[ \Phi_\src = \mathcal{F}_\src + \phi(F_\src,F_\tgt) \]
\[ \Phi_\tgt = \mathcal{F}_\tgt + \phi(F_\tgt,F_\src), \]
where $\phi: \Re^{N\times K} \times \Re^{N\times K} \rightarrow \Re^{N\times K}  $ denotes the function learned by the Transformer.

As both embeddings  $\Phi_\src$ and $\Phi_\tgt$ are in $\Re^{N\times K}$, we derive a global embedding through the maxpool operator. We then use an MLP to extract the reward vector from this embedding. For this, we consider that the rotational and translational actions are completely different in nature, although the form of the reward shares some similarities. We thus make use of a single MLP to learn any shared information followed by separate MLPs for the two types of actions.

\begin{figure}
\centering
\begin{subfigure}[b]{0.49\linewidth}
	\centering
	\includegraphics[width=\linewidth,trim={1.7cm 2cm 1.7cm 2cm},clip]{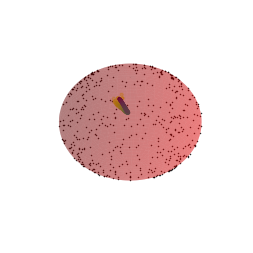}
\caption{}
\end{subfigure}
\begin{subfigure}[b]{0.49\linewidth}
	\centering
	\includegraphics[width=\linewidth,trim={1.7cm 2cm 1.7cm 2cm},clip]{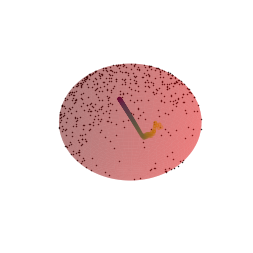}
\caption{}
\end{subfigure}
\caption{Sampled rotations using (a) the isotropic method and (b) the naive method for angles up until $180^\circ$, where the axis of rotation is represented as a point. The highlighted lines are registration paths (evolving from purple to yellow) trained with the respective sampling methods. In (b), the registration path goes to a sparsely sampled region, which can lead to misregistration.}
\label{fig:sampling}
\end{figure}

\subsection{Sampling Rotations from the $\SO(3)$ Manifold}
\label{sec:sampling_diff}
\begin{figure}
\begin{subfigure}[b]{0.32\linewidth}
	\centering
	\includegraphics[width=\linewidth,trim={1cm 0cm 0cm 0cm},clip]{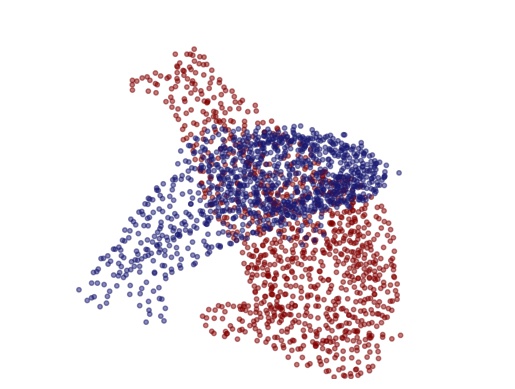}
\caption{Input }
\end{subfigure}
\begin{subfigure}[b]{0.32\linewidth}
	\centering
	\medskip
	\includegraphics[width=\linewidth,trim={0cm 2cm 1cm 1cm},clip]{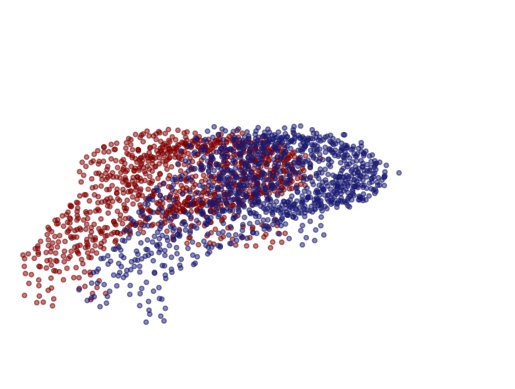}
\label{fig:naive}
\caption{Naive}
\end{subfigure}
\begin{subfigure}[b]{0.32\linewidth}
	\centering
	\includegraphics[width=\linewidth,trim={0cm 2cm 1cm 1cm},clip]{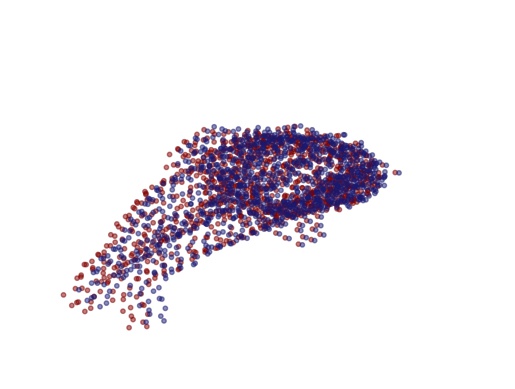}
\label{fig:isotropic}
\caption{Isotropic}
\end{subfigure}
\caption{Example source and target point clouds registered through training by the naive method and the isotropic method.}
\label{fig:sampling_reg}
\end{figure}

When generating the transformations used for the training samples, many learning based methods in the literature obtain the rotations by sampling the Euler angles uniformly at random~\cite{Wang, Yew2020}. This, however, causes a bias in sampling the SO(3) space as shown in Fig.~\ref{fig:sampling}~(b) as the rotations do not adequately cover the manifold. 
We instead sample our transformations uniformly in $\SO(3)$ by sampling the axis of rotation uniformly at random and sampling the angle according to the Haar measure of $\SO(3)$ \cite{Diaconis1987,Kuffner2004}. As shown in Fig.~\ref{fig:sampling}~(a), this approach removes any bias from learned transformations and results in a better coverage of the rotation manifold. A bias in sampling can result in subpar performance for iterative learning-based methods as successive transformations can end up in parts of the manifold that have not been sampled well. This phenomenon is illustrated with the highlighted paths in Fig.~\ref{fig:sampling} for a sample registration problem. When the registration path wanders into the undersampled parts of the manifold, misregistration is likely as the learning based methods might not have seen enough samples from these undersampled parts during training. We show the registration results for such an example in Fig.~\ref{fig:sampling_reg} that resulted in misregistration when the network is trained with the naive sampling method. In Section~\ref{sec:ablation_sampling}, we compare the results of the full test set between the two sampling methods.

\subsection{Loss function}
The network outputs a reward vector $h \in \Re^{|A|}$, where each element of the vector estimates the reward for one of the actions in $A$. We use a straight-forward loss function to measure the agreement of this reward vector with the ground truth $g \in \Re^{|A|}$ and include a $\ell_2$ weight decay with constant $\lambda$ and network weights $\bm{\Theta}$:
\[ \mathrm{Loss} = \frac{1}{|A|} || g - h||^2 + \lambda ||\bm{ \Theta}||^2  \]

\begin{figure*}[t]
\centering
\begin{subfigure}[b]{0.19\linewidth}
	\centering
	\includegraphics[width=\linewidth,trim={2cm 2cm 2cm 2cm},clip]{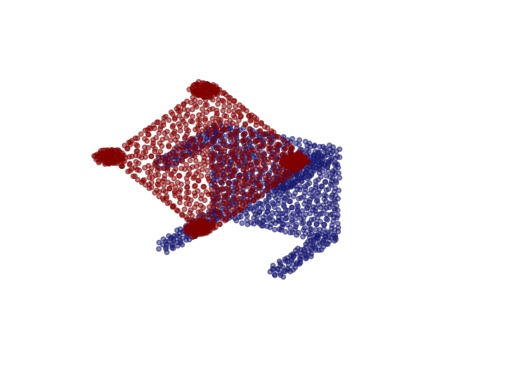}
	\includegraphics[width=\linewidth,trim={2cm 2cm 2cm 2cm},clip]{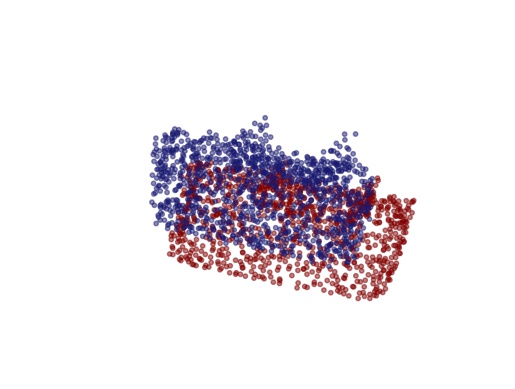}
	\includegraphics[width=\linewidth,trim={2cm 2cm 2cm 2cm},clip]{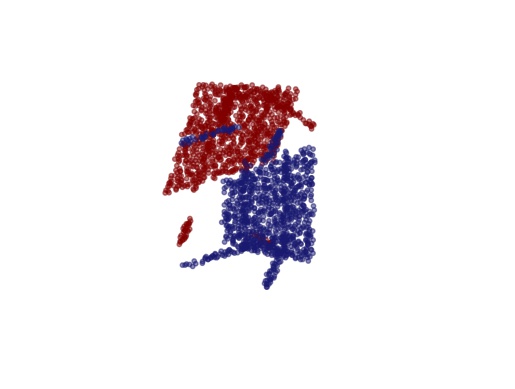}
	\caption{\label{fig:fig1} 0 iterations}
\end{subfigure}
\begin{subfigure}[b]{0.19\linewidth}
	\centering
	\includegraphics[width=\linewidth,trim={2cm 2cm 2cm 2cm},clip]{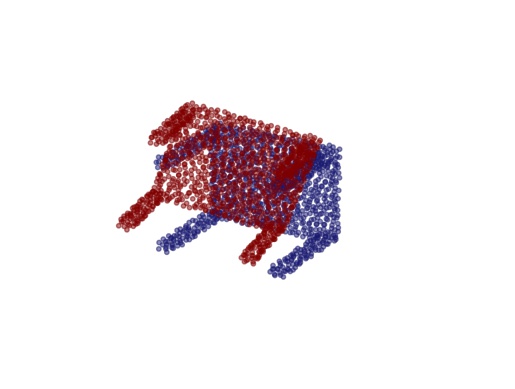}
	\includegraphics[width=\linewidth,trim={2cm 2cm 2cm 2cm},clip]{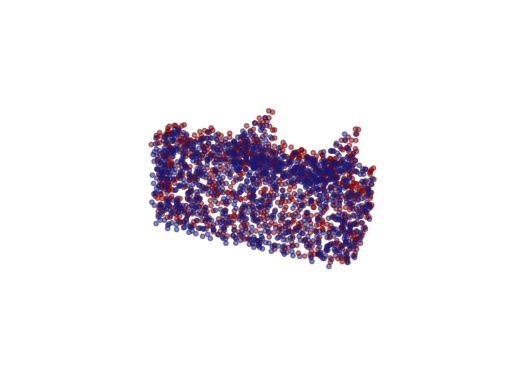}
	\includegraphics[width=\linewidth,trim={2cm 2cm 2cm 2cm},clip]{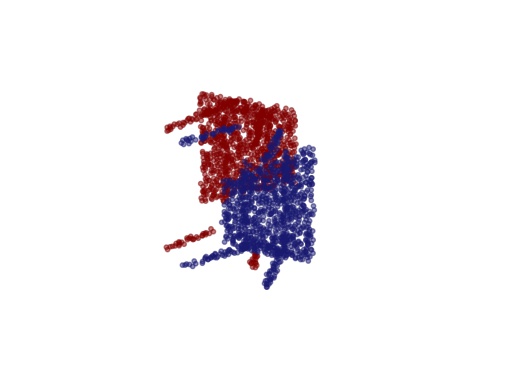}
	\caption{\label{fig:fig2} 10 iterations}
\end{subfigure}
\begin{subfigure}[b]{0.19\linewidth}
	\centering
	\includegraphics[width=\linewidth,trim={2cm 2cm 2cm 2cm},clip]{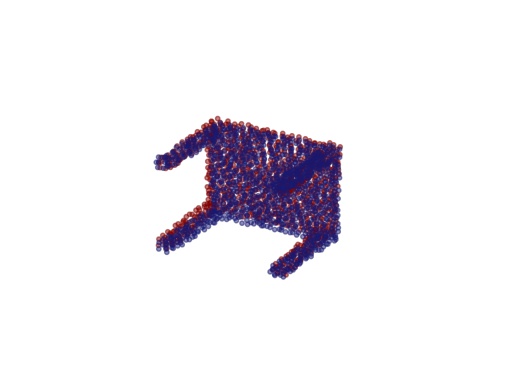}
		\includegraphics[width=\linewidth,trim={2cm 2cm 2cm 2cm},clip]{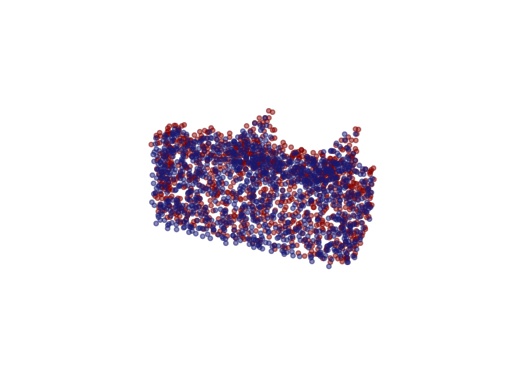}
		\includegraphics[width=\linewidth,trim={2cm 2cm 2cm 2cm},clip]{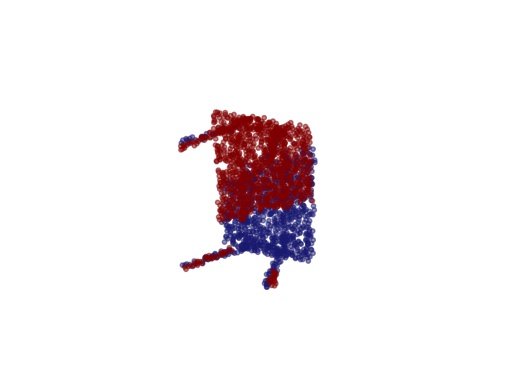}
	\caption{\label{fig:fig3} 20 iterations}
\end{subfigure}
\begin{subfigure}[b]{0.19\linewidth}
	\centering
	\includegraphics[width=\linewidth,trim={2cm 2cm 2cm 2cm},clip]{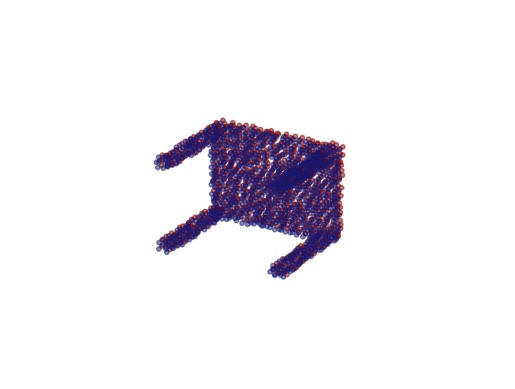}
		\includegraphics[width=\linewidth,trim={2cm 2cm 2cm 2cm},clip]{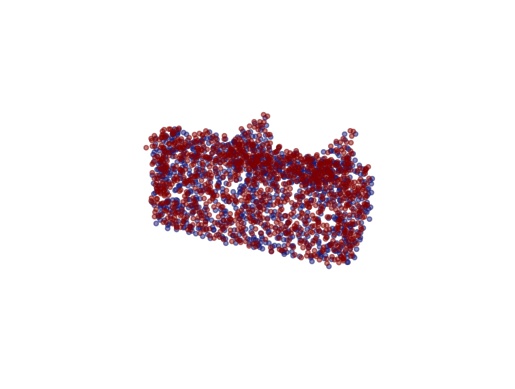}
		\includegraphics[width=\linewidth,trim={2cm 2cm 2cm 2cm},clip]{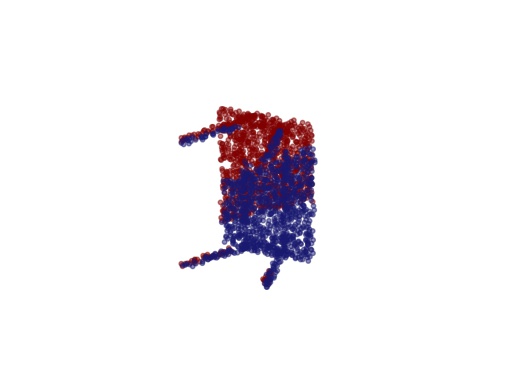}
	\caption{\label{fig:fig4} 40 iterations}
\end{subfigure}
\begin{subfigure}[b]{0.19\linewidth}
	\centering
	\includegraphics[width=\linewidth,trim={2cm 2cm 2cm 2cm},clip]{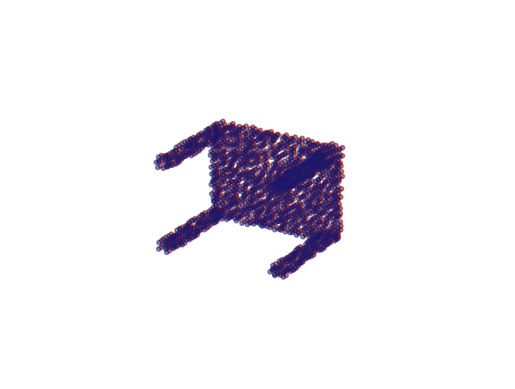}
		\includegraphics[width=\linewidth,trim={2cm 2cm 2cm 2cm},clip]{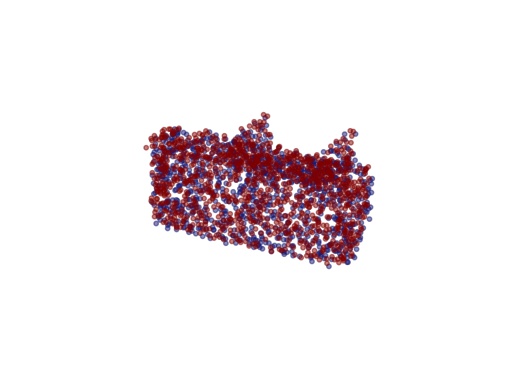}
		\includegraphics[width=\linewidth,trim={2cm 2cm 2cm 2cm},clip]{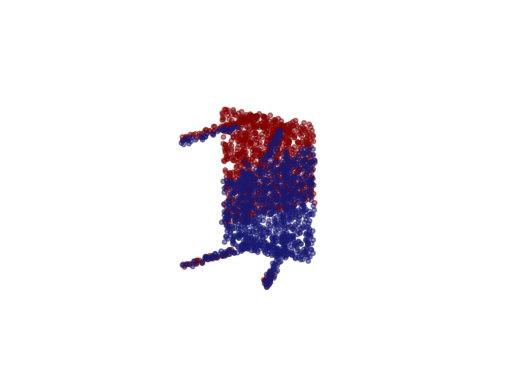}
	\caption{\label{fig:fig5} 60 iterations}
\end{subfigure}

\caption{Example registration processes for clean, noisy and partially visible examples using our iterative method.}
\label{fig:iterative_registration}
\end{figure*}

\section{Experiments}

The experiments are conducted using the ModelNet40 dataset~\cite{Wu2014}. We split the official training set into distinct training and validation sets, and retain the official test set. We train and validate on the first 20 categories, then test on the latter 20. This gives us 4,603 training samples, 509 validation samples and 1,266 testing samples.

Since our method uses discrete steps which places a limit on the performance, we apply ICP on top of the result. We call this method PRANet-v2. The method without ICP is denoted PRANet-v1. 

\subsubsection{Benchmarks and metrics.} We benchmark our approach against both learning-based and classical methods. For the former category, we chose DCP-v2~\cite{Wang} and RPMNet~\cite{Yew2020}, whereas for the latter we chose ICP and FGR. We use the Open3D 0.9.0 implementations of ICP and FGR, and adapt the released code for DCP-v2 and RPMNet for our experiments. For both RPMNet and FGR, the point cloud features were enhanced with surface normals, whereas the other methods, including ours, only had positional data as input. 

In terms of the evaluation metrics, we use the isotropic rotation error and translation error, the $\ell_2$ norm of the complete and clean point clouds and the modified Chamfer distance as proposed by~\cite{Yew2020}. For all these metrics, a lower value indicates a better registration result.

\subsubsection{Parameters.} The architecture details of our network is as follows. We have 5 EdgeConv layers in the DGCNN component, where the number of filters in each layer are [64, 64, 128, 256, 1024]. There are 4 heads in the multi-head attention in the Transformer component, and an embedding dimension of 1024. We use LayerNorm without Dropout. We use the SGD optimizer with an initial learning rate of $0.1$, which is then decayed at epochs 300, 700 and 1000 by a factor of 10. We use a weight decay constant of $10^{-4}$. The network is implemented using PyTorch 1.1.0. We set the random seed as 1234.

\subsection{Clean Dataset}

The first experiment consists of the ModelNet40 dataset with no noise or occlusions added. Following the methodology of~\cite{Qi2017} and~\cite{Wang}, we only use the $(x,y,z)$ positional features of the point cloud and randomly sample 1024 points from the surface of each object. The point cloud is then centered at the origin and normalized to fit in the unit sphere. As for the transformation, we sample the rotation uniformly at random according to the $\SO(3)$ geometry up until $60^\circ$ along any axis, and the translation is uniformly sampled between $[-0.5,0.5]$. To enhance training, we make use of curriculum learning, where the network starts with rotation angles in $[-10^\circ,10^\circ]$ and translation magnitudes up to $0.5/7$ for each axis at epochs 0-69, and continues with the full range of transformations for the remaining epochs, up until a total of 1300 epochs. For the actions, we use fixed angles of $\pm0.5^\circ$  and $\pm10^\circ$ around each of the three axes and translations of $\pm 0.01$ and $\pm0.1$ in each of the three axis directions. This gives a total of 24 actions. The corresponding rewards for each action step size are normalized separately to have unit $\ell_2$ norm. At test time, the registration is allowed to run for 60 iterations. For the first 20 iterations, the rewards for only the larger actions $(\pm 10^\circ, \pm0.1)$ are considered and only these actions are applied. For the remaining 40 iterations only the smaller actions $(\pm 0.5^\circ, \pm 0.01)$ and their corresponding rewards are used. Unless otherwise stated, this will be our default configuration in all subsequent experiments. Table~\ref{table:clean} shows state-of-the art performance from PRANet-v2 compared to classical and learning-based methods.
\begin{table}
    \centering
    \begin{tabular}{@{}l l l l l l l@{}}
    \toprule Model & Rot. & Trans.  & L2 & MCD \\ \midrule
    ICP& 5.872&0.0463  &0.0507  & 0.00215 \\
    FGR*& 0.052& 0.000407 & 0.000342 & 0.000107 \\ 
    RPMNet*& 0.061 & 0.000334 &  0.000184 & 0.0000053\\
    DCP-v2& 3.441& 0.0250   & 0.0283  & 0.00157 \\
    \midrule
    PRANet-v1 & 1.554 & 0.0246 & 0.0249 & 0.00131 \\
    PRANet-v2& 0.045  & 0.000471 & 0.000254 & 0.0000043 \\
    \bottomrule
   \end{tabular}
   \caption{Clean dataset. The asterisks mark the methods that use input features enhanced with surface normals. The metrics shown, from left to right, are the isotropic rotation and translation errors, the $\ell_2$ distance of the clean point clouds and the modified Chamfer distance.}
   \label{table:clean}
\end{table}

\begin{table}[t]
    \centering
    \begin{tabular}{@{}l l l l l@{}}
    \toprule Model & Rot & Trans  & L2 & MCD \\ \midrule
    ICP& 	6.393 & 0.0474  & 0.0517 &  0.00221\\
    FGR* & 3.879 & 0.0307  & 0.0322  & 0.00270 \\
    RPMNet* & 0.637 & 0.00769&  0.00786 & 0.000672 \\
    DCP-v2& 11.723 & 0.0869 & 0.0922 & 0.00482 \\
    \midrule
    PRANet-v1 & 5.083 & 0.0448 & 0.0457 & 0.00183\\
    PRANet-v2 & 2.900  & 0.0207 & 0.0213 & 0.000768 \\
    \bottomrule
   \end{tabular}
   \caption{Noisy dataset.}
   \label{table:noisy}
\end{table}
\begin{table}[t]
    \centering
    \begin{tabular}{@{}l l l l l@{}}
    \toprule Model & Rot & Trans  & L2 & MCD \\ \midrule
    ICP& 24.692&0.251  & 0.270  & 0.0134 \\
    FGR*& 73.734 & 0.477  & 0.540  & 0.0447 \\
    RPMNet* & 3.514 & 0.0388 & 0.0400 & 0.00161 \\
        DCP-v2& 21.318 & 0.204 & 0.228 & 0.0155\\ \midrule
    PRANet-v1 & 8.902 & 0.119 & 0.124 & 0.00625 \\
    PRANet-v2 & 7.237 & 0.0759 & 0.0785 & 0.00239 \\
    \bottomrule
   \end{tabular}
   \caption{Partial visibility.}
   \label{table:partial}
\end{table}
\begin{table}[t]
    \centering
    \begin{tabular}{@{}l l@{}}
    \toprule Model & Time (ms) \\ \midrule
    ICP & 131 \\
    FGR & 157 \\
    RPMNet & 173 \\
    DCP-v2& 24	\\ \midrule
    PRANet-v1 & 742  \\
    \bottomrule
   \end{tabular}
   \caption{Average inference time.}
   \label{table:inference_time}
\end{table}

 

\begin{table}[t]
    \centering
    \begin{tabular}{@{}l l l l l l l@{}}
    \toprule Distance in Reward & Rot & Trans  & L2 & MCD \\ \midrule
    \textbf{SE(3)} &  1.554 & 0.0246 & 0.0249 & 0.00131 \\
    $\ell_2$ Distance & 6.090 & 0.0480 & 0.0456 & 0.00189	\\
    MCD & 6.034 & 0.0577 & 0.0588 & 0.00219	\\
    \bottomrule
   \end{tabular}
   \caption{Reward function. The method in bold is PRANet-v1.}
      \label{table:rewards}
\end{table}

\begin{table}[t]
    \centering
    \begin{tabular}{@{}l l l l l l@{}}
    \toprule Method & Rot & Trans  & L2 & MCD \\ \midrule
    \textbf{Isotropic} & 1.555 & 0.0240 & 0.0247 & 0.00129  \\
    Naive & 2.771 & 0.0302 & 0.0307 & 0.00128 \\
    \bottomrule
   \end{tabular}
   \caption{Sampling rotations.}
   \label{table:sampling}
\end{table}

\begin{table}[t]
    \centering
    \begin{tabular}{@{}l l l l l l@{}}
    \toprule Method & Rot & Trans  & L2 & MCD \\ \midrule
    \textbf{Curriculum} & 1.554 & 0.0246 & 0.0249 & 0.00131  \\
    Uniform & 4.611 & 0.0399 & 0.0400 & 0.00179 \\
    Ad-hoc & 3.374 & 0.0289 & 0.0284 & 0.00074 \\
    \bottomrule
   \end{tabular}
   \caption{Curriculum learning.}
   \label{table:curriculum}
\end{table}

\begin{table}[t]
    \centering
    \begin{tabular}{@{}l l l l l l@{}}
    \toprule Policy & Rot & Trans  & L2 & MCD \\ \midrule
    \textbf{deterministic}  & 1.554 & 0.0246 & 0.0249 & 0.00131 \\ 
    stoch1 &1.550 & 0.0248 & 0.0250 & 0.00132 \\
    stoch2  & 3.669 &  0.0544 &  0.0556 & 0.00484 \\
    \bottomrule
   \end{tabular}
   \caption{Stochasticity of the policy.}
   \label{table:policy}
\end{table}

\subsection{Gaussian Noise}
In this experiment, we evaluate our method in the presence of noisy data with sampling differences bringing us closer to real world scenarios. Using a similar methodology to that in~\cite{Yew2020}, we sample 1024 points from the source and target point cloud independently, and then add noise sampled from $\mathcal{N} (0,0.01)$, which is clipped to $\pm 0.05$ for each axis. Table~\ref{table:noisy} shows a summary of the results. Note that the normal data for both RPMNet and FGR are clean and had no noise added, allowing these methods to perform particularly well. PRANet-v2 is the second best performing method, while PRANet-v1 performs comparably with FGR, showing a superior performance in terms of the modified Chamfer distance.
 
\subsection{Partial Visibility}
We further increase the difficulty by considering partially visible point clouds, while retaining the Gaussian noise and the sampling differences described in the previous section. Such deviations occur frequently in real world point cloud data. As is done in~\cite{Yew2020}, we sample a random plane that passes through the origin, translate it along its normal vector, and then retain 70\% of the points. The point cloud is also downsampled to 717 points to maintain a similar point density to that in previous experiments. Similar to the previous section, no noise was added to the normals, which are additional input features for RPMNet and FGR. Table~\ref{table:partial} shows a summary of the results. Both versions of PRANet outperform all the other approaches apart from RPMNet.

\subsection{Computational Efficiency}
Table~\ref{table:inference_time} shows the inference time of our method as compared with the other approaches. These experiments were done on an  Intel Xeon(R) CPU E5-2650 v3 @ 2.30GHz machine with 1 NVIDIA GeForce GTX Titan X GPU and 25GB memory. We estimate the average time by computing the time taken on all 1,266 samples of the unseen categories and averaging. Note that the ICP and FGR implementations run entirely on the CPU. As PRANet is iterative, it requires multiple queries to the artificial agent and hence is more computationally costly than the other methods.

\subsection{Agent Design}
We conduct various experiments to justify our setup. All experiments are done without the addition of ICP.

\subsubsection{Number of iterations.}
\begin{figure}[H]
    \centering
    \includegraphics[width=\linewidth]{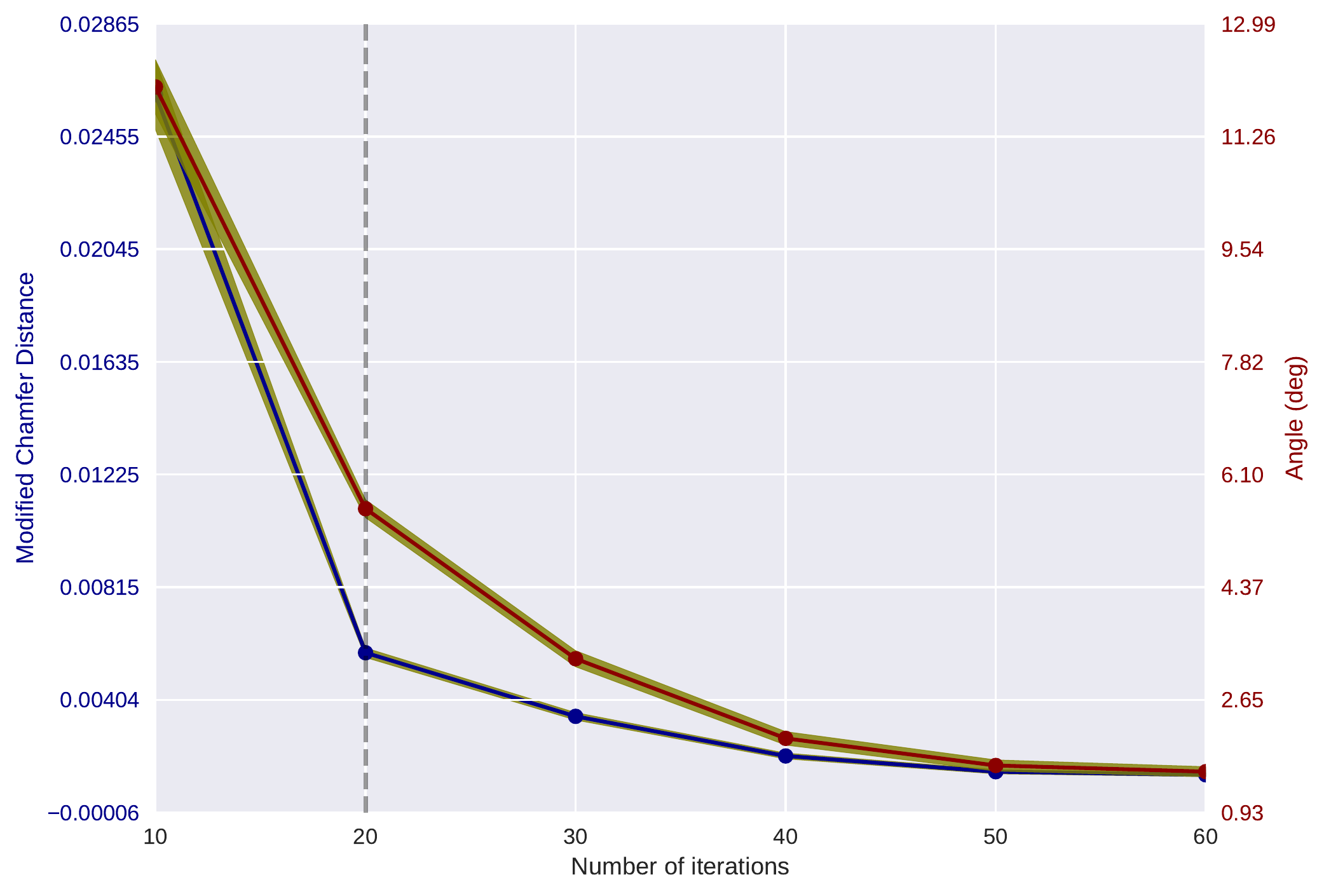}
    \caption{The Chamfer distance and the angle over the number of iterations. The initial Chamfer distance is 0.199 and the initial angle is $43.7^\circ$, which are omitted from the plot for clarity. The yellow region marks the 95\% confidence interval, and the vertical dotted line indicates the change in action step sizes.}
    \label{fig:cd_numit}
\end{figure}

Fig.~\ref{fig:cd_numit} shows the Chamfer distance and the isotropic rotation error (the angle) over the number of iterations, averaged over the entire test set. Fig.~\ref{fig:iterative_registration} gives some qualitative results. From both figures, it can be seen that the performance gain is maximal towards the start of the registration process, and then gradually decreases as the registration progresses.

\subsubsection{Reward function.}
Our reward function is based on the distances of the $\SE(3)$ transformations. Instead of using the distance of the transformation, an alternative would be to use point-based distances. We formulate distances in terms of the $\ell_2$ distance of the clean and complete point clouds as well as the modified Chamfer distance.  In particular, for each action $a \in A$, \[r_a(\src,\tgt) = D(\src,\tgt) - D(\src^{\prime}_{a}, \tgt)\] where $\src^{\prime}_{a}$ is the transformed source point cloud using action $a$ and $ D$ is the distance in question. Similar to the transformation distance case, the rotation actions are disentangled from the translations by rotating around the centroid of the observed source point cloud.
Table~\ref{table:rewards} shows the performance of these alternate reward functions. The reward based on the $\SE(3)$ transformation distance outperforms both point-based reward functions.

\subsubsection{Sampling rotations.}
\label{sec:ablation_sampling}
To compare the results for the two sampling methods as introduced in Section~\ref{sec:sampling_diff}, we create another test set that samples the transformations according to the naive method, up to a maximum rotation of $32^\circ$ around each axis. This corresponds to a maximum angle of approximately $60^\circ$ in the axis-angle representation, although the larger angles are only achieved for a small set of rotation axes. For the naive method, the transformations are sampled exactly in this range at training time. As for the isotropic sampling method, the transformations are sampled up to a maximum angle of $60^\circ$, where the larger angles can be sampled for any rotation axis. Table~\ref{table:sampling} gives the results, showing that the isotropic method achieves an improved performance over the naive method.

\subsubsection{Curriculum learning.}
We applied curriculum learning to sample the transformations of the point clouds during training time. This ensures that rewards are learned explicitly when transformations are small at the initial phase of the training. To test whether the use of curriculum learning is critical to the success of our method, we evaluate this against a uniform sampling schedule and an ad-hoc sampling method, where small and large transformations are each sampled with a probability of 50\%. This comparison is shown in Table~\ref{table:curriculum}. The sampling method with curriculum learning outperforms the other two methods in all metrics apart from the MCD, where the ad-hoc method does better. 

\subsubsection{Stochasticity of the policy.}
Our current policy is entirely deterministic. Given the vector of rewards, we always pick the action with the largest reward. We evaluate whether a stochastic policy, which could potentially help with local minima, would give better results. The first modification that we make is to have the top three actions picked with probability 0.85, 0.15, and 0.05 respectively. We denote this modification ``stoch1". Another variant that we consider is where all actions with positive rewards are considered equally, and are attributed with equal probabilities. This is motivated by the fact that each such action would bring a reduction in the distance. In the event that none of the actions have a positive reward, the action with the top reward is picked. We denote this modification ``stoch2". Table~\ref{table:policy} gives a comparison of these policy choices. The ``stoch1" policy gives similar results to the deterministic policy, while the other stochastic policy leads to a worse performance.

\section{Conclusion}
We propose a greedy approach to train an artificial agent by learning rewards using deep supervised learning, and show promising results in experiments with the ModelNet40 dataset with clean, noisy and partially visible data. Unlike the prevailing learning-based methods, our approach does not rely on point matching, and frames the point cloud registration problem as a Markov Decision Process. 

Open problems include possible additions and modifications that can enhance the performance. For instance, we used a generic point cloud embedding that was designed for point cloud matching. It could be possible to design a more specific embedding for learning rewards. We also use the maxpool operator to obtain a global embedding, from which we derive the reward vector. An improved architecture would take local information into account as well to obtain the final reward vector.
In addition, for simplicity, we used a greedy approach, where we only take the action that maximizes the reward of the next action. An extension could include taking the action that maximizes the cumulative reward of the next $n$ steps, with a possible discount factor.


\pagebreak
\bibliography{pointreg}

\begin{thebibliography}{21}
\providecommand{\natexlab}[1]{#1}

\bibitem[{Aoki et~al.(2019)Aoki, Goforth, Srivatsan, and Lucey}]{Aoki2019}
Aoki, Y.; Goforth, H.; Srivatsan, R.~A.; and Lucey, S. 2019.
\newblock {PointNetLK: Robust \& Efficient Point Cloud Registration using
  PointNet}.
\newblock \emph{Conference on Computer Vision and Pattern Recognition}.

\bibitem[{Bauer, Patten, and Vincze(2021)}]{Bauer2021}
Bauer, D.; Patten, T.; and Vincze, M. 2021.
\newblock {ReAgent: Point Cloud Registration using Imitation and Reinforcement
  Learning}.
\newblock \emph{Conference on Computer Vision and Pattern Recognition}.

\bibitem[{Diaconis and Shahshahani(1987)}]{Diaconis1987}
Diaconis, P.; and Shahshahani, M. 1987.
\newblock {The Subgroup Algorithm for Generating Uniform Random Variables}.
\newblock \emph{Probability in the Engineering and Informational Sciences},
  1(1): 15--32.

\bibitem[{Fitzgibbon(2003)}]{Fitzgibbon2003}
Fitzgibbon, A.~W. 2003.
\newblock {Robust registration of 2D and 3D point sets}.
\newblock In \emph{Image and Vision Computing}, volume~21, 1145--1153.

\bibitem[{Fu et~al.(2021)Fu, Liu, Luo, and Wang}]{rgmnet}
Fu, K.; Liu, S.; Luo, X.; and Wang, M. 2021.
\newblock {Robust Point Cloud Registration Framework Based on Deep Graph
  Matching}.
\newblock \emph{Conference on Computer Vision and Pattern Recognition}.

\bibitem[{Huynh(2009)}]{Huynh2009}
Huynh, D.~Q. 2009.
\newblock {Metrics for 3D rotations: Comparison and analysis}.
\newblock \emph{Journal of Mathematical Imaging and Vision}, 35(2): 155--164.

\bibitem[{Kuffner(2004)}]{Kuffner2004}
Kuffner, J.~J. 2004.
\newblock {Effective sampling and distance metrics for 3D rigid body path
  planning}.
\newblock In \emph{Proceedings - IEEE International Conference on Robotics and
  Automation}, volume 2004, 3993--3998.

\bibitem[{Li et~al.(2020)Li, Zhang, Xu, Zhou, and Zhang}]{idam}
Li, J.; Zhang, C.; Xu, Z.; Zhou, H.; and Zhang, C. 2020.
\newblock {Iterative Distance-Aware Similarity Matrix Convolution with
  Mutual-Supervised Point Elimination for Efficient Point Cloud Registration}.
\newblock \emph{European Conference on Computer Vision}.

\bibitem[{Liao et~al.(2017)Liao, Miao, {De Tournemire}, Grbic, Kamen, Mansi,
  and Comaniciu}]{Miao2017}
Liao, R.; Miao, S.; {De Tournemire}, P.; Grbic, S.; Kamen, A.; Mansi, T.; and
  Comaniciu, D. 2017.
\newblock {An artificial agent for robust image registration}.
\newblock In \emph{AAAI Conference on Artificial Intelligence}, volume~31,
  4168--4175.

\bibitem[{Lucas and Kanade(1981)}]{Lucas1981}
Lucas, B.~D.; and Kanade, T. 1981.
\newblock { An iterative image registration technique with an application to
  stereo vision.}
\newblock In \emph{International Joint Conference on Artificial Intelligence}.

\bibitem[{Qi et~al.(2017)Qi, Su, Mo, and Guibas}]{Qi2017}
Qi, C.~R.; Su, H.; Mo, K.; and Guibas, L.~J. 2017.
\newblock {PointNet: Deep learning on point sets for 3D classification and
  segmentation}.
\newblock \emph{Conference on Computer Vision and Pattern Recognition}, 77--85.

\bibitem[{Rusinkiewicz and Levoy(2001)}]{Rusinkiewicz2001}
Rusinkiewicz, S.; and Levoy, M. 2001.
\newblock {Efficient variants of the ICP algorithm}.
\newblock \emph{Proceedings of International Conference on 3-D Digital Imaging
  and Modeling, 3DIM}, 145--152.

\bibitem[{Segal, Haehnel, and Thrun(2009)}]{Segal}
Segal, A.~V.; Haehnel, D.; and Thrun, S. 2009.
\newblock {Generalized-ICP}.
\newblock \emph{Robotics: Science and Systems}, 2: 435.

\bibitem[{Wang and Solomon(2019{\natexlab{a}})}]{Wang2019}
Wang, Y.; and Solomon, J. 2019{\natexlab{a}}.
\newblock {PRNet: Self-supervised learning for partial-to-partial
  registration}.
\newblock In \emph{Advances in Neural Information Processing Systems},
  volume~32.

\bibitem[{Wang and Solomon(2019{\natexlab{b}})}]{Wang}
Wang, Y.; and Solomon, J.~M. 2019{\natexlab{b}}.
\newblock {Deep Closest Point: Learning Representations for Point Cloud
  Registration}.
\newblock In \emph{International Conference on Computer Vision}.

\bibitem[{Wang et~al.(2019)Wang, Sun, Liu, Sarma, Bronstein, and
  Solomon}]{DGCNN}
Wang, Y.; Sun, Y.; Liu, Z.; Sarma, S.~E.; Bronstein, M.~M.; and Solomon, J.~M.
  2019.
\newblock {Dynamic Graph CNN for Learning on Point Clouds}.
\newblock \emph{ACM Transactions on Graphics}, 38(5): 13.

\bibitem[{Wu et~al.(2014)Wu, Song, Khosla, Yu, Zhang, Tang, and Xiao}]{Wu2014}
Wu, Z.; Song, S.; Khosla, A.; Yu, F.; Zhang, L.; Tang, X.; and Xiao, J. 2014.
\newblock {3D ShapeNets: A Deep Representation for Volumetric Shapes}.
\newblock In \emph{Conference on Computer Vision and Pattern Recognition}.

\bibitem[{Yang, Shi, and Carlone(2021)}]{Yang2021}
Yang, H.; Shi, J.; and Carlone, L. 2021.
\newblock {Teaser: Fast and certifiable point cloud registration}.
\newblock \emph{IEEE Transactions on Robotics}, 37(2): 314--333.

\bibitem[{Yang et~al.(2016)Yang, Li, Campbell, and Jia}]{Yang2016}
Yang, J.; Li, H.; Campbell, D.; and Jia, Y. 2016.
\newblock {Go-ICP: A Globally Optimal Solution to 3D ICP Point-Set
  Registration}.
\newblock \emph{IEEE Transactions on Pattern Analysis and Machine
  Intelligence}, 38(11): 2241--2254.

\bibitem[{Yew and Lee(2020)}]{Yew2020}
Yew, Z.~J.; and Lee, G.~H. 2020.
\newblock {RPM-Net: Robust point matching using learned features}.
\newblock In \emph{Conference on Computer Vision and Pattern Recognition}.

\bibitem[{Zhou, Park, and Koltun(2016)}]{Zhou2016}
Zhou, Q.-Y.; Park, J.; and Koltun, V. 2016.
\newblock {Fast Global Registration}.
\newblock In \emph{European Conference on Computer Vision}.

\end{thebibliography}


\end{document}